\documentclass[sigconf]{acmart}

\usepackage{float}
\usepackage{subcaption}
\usepackage{multirow}
\usepackage[normalem]{ulem}

\AtBeginDocument{%
  \providecommand\BibTeX{{%
    \normalfont B\kern-0.5em{\scshape i\kern-0.25em b}\kern-0.8em\TeX}}}

\copyrightyear{2022}
\acmYear{2022}
\setcopyright{acmcopyright}
\acmConference[KDD '22] {Proceedings of the 28th ACM SIGKDD Conference on Knowledge Discovery and Data Mining}{August 14--18, 2022}{Washington, DC, USA.}
\acmBooktitle{Proceedings of the 28th ACM SIGKDD Conference on Knowledge Discovery and Data Mining (KDD '22), August 14--18, 2022, Washington, DC, USA}
\acmPrice{15.00}
\acmISBN{978-1-4503-9385-0/22/08}
\acmDOI{10.1145/3534678.3539448}
\begin{document}

\title{Robust Inverse Framework using Knowledge-guided Self-Supervised Learning: An application to Hydrology}

\author{Rahul Ghosh}
\affiliation{%
  \institution{University of Minnesota - Twin Cities}
  \city{Minneapolis}
  \state{Minnesota}
  \country{USA}
}
\email{ghosh128@umn.edu}

\author{Arvind Renganathan}
\affiliation{%
  \institution{University of Minnesota - Twin Cities}
  \city{Minneapolis}
  \state{Minnesota}
  \country{USA}
}
\email{renga016@umn.edu}

\author{Kshitij Tayal}
\affiliation{%
  \institution{University of Minnesota - Twin Cities}
  \city{Minneapolis}
  \state{Minnesota}
  \country{USA}
}
\email{tayal007@umn.edu}

\author{Xiang Li}
\affiliation{%
  \institution{University of Minnesota - Twin Cities}
  \city{Minneapolis}
  \state{Minnesota}
  \country{USA}
}
\email{lixx5000@umn.edu}

\author{Ankush Khandelwal}
\affiliation{%
  \institution{University of Minnesota - Twin Cities}
  \city{Minneapolis}
  \state{Minnesota}
  \country{USA}
}
\email{khand035@umn.edu}

\author{Xiaowei Jia}
\affiliation{%
  \institution{University of Pittsburgh}
  \city{Pittsburgh}
  \state{Pennsylvania}
  \country{USA}
}
\email{xiaowei@pitt.edu}

\author{Christopher Duffy}
\affiliation{%
  \institution{Penn State University}
  \city{State College}
  \state{Pennsylvania}
  \country{USA}
}
\email{cxd11@psu.edu}

\author{John Nieber}
\affiliation{%
  \institution{University of Minnesota - Twin Cities}
  \city{Minneapolis}
  \state{Minnesota}
  \country{USA}
}
\email{nieber@umn.edu}

\author{Vipin Kumar}
\affiliation{%
  \institution{University of Minnesota - Twin Cities}
  \city{Minneapolis}
  \state{Minnesota}
  \country{USA}
}
\email{kumar001@umn.edu}

\renewcommand{\shortauthors}{Ghosh, et al.}

\begin{abstract}
Machine Learning is beginning to provide state-of-the-art performance in a range of environmental applications such as streamflow prediction in a hydrologic basin. However, building accurate broad-scale models for streamflow remains challenging in practice due to the variability in the dominant hydrologic processes, which are best captured by sets of process-related basin characteristics. Existing basin characteristics suffer from noise and uncertainty, among many other things, which adversely impact model performance. To tackle the above challenges, in this paper, we propose a novel Knowledge-guided Self-Supervised Learning (KGSSL) inverse framework to extract system characteristics from driver and response data. This first-of-its-kind framework achieves robust performance even when characteristics are corrupted. We show that KGSSL achieves state-of-the-art results for streamflow modeling for CAMELS (Catchment Attributes and MEteorology for Large-sample Studies) which is a widely used hydrology benchmark dataset. Specifically, KGSSL outperforms other methods by up to 16 \% in reconstructing characteristics. Furthermore, we show that KGSSL is relatively more robust to distortion than baseline methods, and outperforms the baseline model by 35\% when plugging in KGSSL inferred characteristics.
\end{abstract}



\maketitle
\section{Introduction}
\label{Sec:Introduction}
\begin{figure}[t]
    \centering
    \begin{subfigure}{0.5\linewidth}
        \centering
        \includegraphics[width=\linewidth]{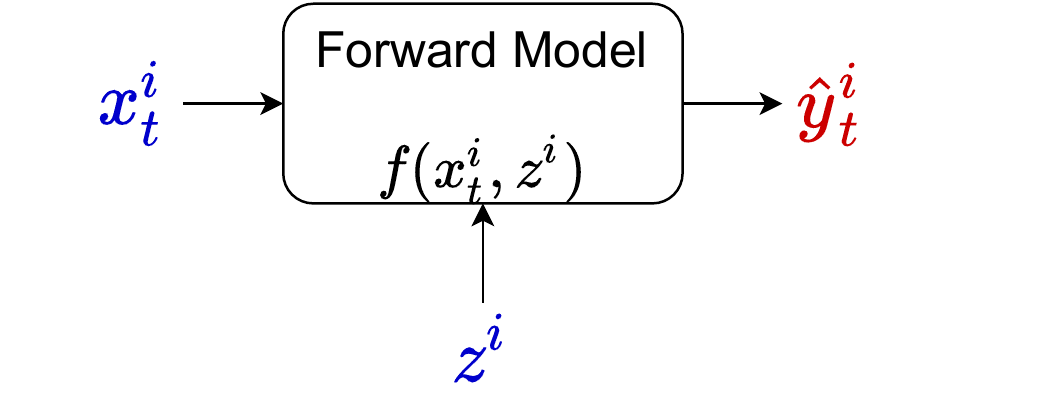}
        \caption{}
        \label{fig:Forward}
    \end{subfigure}%
    \begin{subfigure}{0.5\linewidth}
        \centering
        \includegraphics[width=\linewidth]{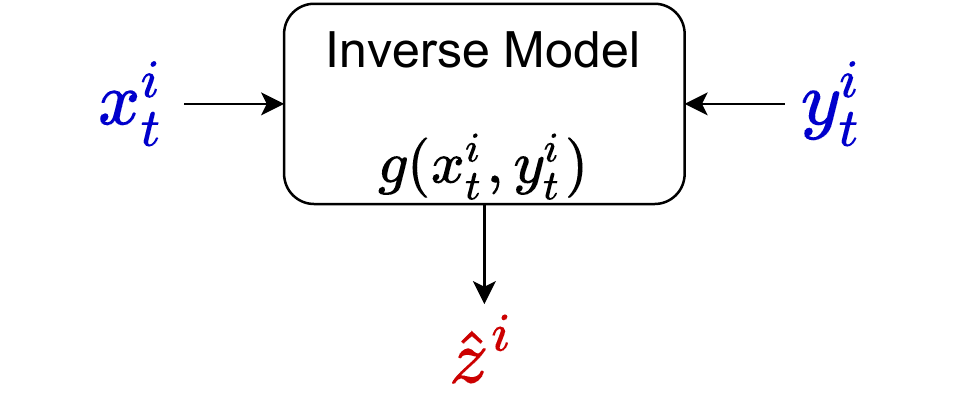}
        \caption{}
        \label{fig:Inverse}
    \end{subfigure}
    \vspace{-.2in}
    \caption{(a) Forward model which uses meteorological drivers $(\boldsymbol{X_i^t})$ and entity characteristics $(\boldsymbol{z_i})$ to predict response $(\boldsymbol{y_i^t})$ (b) The inverse model which approximates entity characteristics $(\boldsymbol{z_i})$ by inverting the forward process.}
    \label{fig:test}
    \vspace{-.2in}
\end{figure}

Machine learning (ML) is increasingly being used to solve challenging tasks in scientific applications such as hydrology, lake sciences and crop yield monitoring. Consider the case of hydrology, where streamflow prediction is one important research problem for understanding hydrology cycles, water supply management, flood mapping, and other operational decisions such as reservoir release. For a given entity (basin/catchment, we use either term interchangeably), the response (streamflow) is governed by drivers (meteorological data e.g., air temperature, precipitation, wind speed) and complex physical processes specific to each entity~\cite{newman2006ecohydrology,bhatt2014tightly}. These complex physical processes are best captured by the inherent characteristics of each entity (e.g., slope, land-cover). For example, for the same amount of precipitation, two basins will have very different streamflow response values depending on their land-cover type. Such problems are often solved using a mechanistic forward model $f$ that predicts the response $y_t$, given drivers $\boldsymbol{x_t}$ and entity characteristic $\boldsymbol{z}$. A wide variety of scientific models can be considered as a mapping function between drivers $\boldsymbol{x_t}$ (e.g. weather drivers, climate forcings), and response $y_t$ (e.g., streamflow in river basin, global average temperature), governed by entity characteristics. Figure~\ref{fig:Forward} shows the diagrammatic representation of this forward model. More recently, Machine Learning (ML) models (e.g. LSTMs) are able to provide state of the art performance for many scientific applications~\cite{kratzert2019towards}. The reason is that ML models are able to benefit from training data from a large cross-section of diverse training data and thus can transfer knowledge across basins. These ML models also learn the mapping from the meteorological drivers and time-invariant basin characteristics to streamflow and thus essentially emulate the forward models~\cite{razavi2013streamflow}.

In the scientific application of streamflow modeling, these entity (basin/catchment) characteristics are only surrogate variables of the true basin characteristics~\cite{beven2020deep} and thus can lead to several challenges. First, there often exists high uncertainty in hydrological measurement, which in turn causes corruption in basin characteristics. Uncertainty can arise due to temporal change, spatial heterogeneity, sufficiency of the characteristic itself to explain the rainfall/runoff process, measurement error, missing data, and correlation among characteristics that collectively contribute to streamflow. Second, the full set of basin characteristics may not be measured across all the river basins, resulting in the incompleteness of basin characteristics. Missing characteristics hinder the building of a global model that can leverage data across multiple basins and constrains the transferability of models built from one region to another. Finally, some basin characteristics may be essential in modeling the rainfall-runoff response relation but may be completely unknown, not well understood, or not present in the available set of basin characteristics. Thus, the ability to infer these time-invariant basin characteristics from the time-varying meteorological and streamflow data is essential for model prediction and hydrological process understanding.

This paper presents an inverse modeling methodology that can be used to identify or reconstruct static characteristics of an environmental system given its input and output over time. Figure ~\ref{fig:Inverse} shows the diagrammatic representation of this inverse problem. Inverse problems \cite{ongie2020deep} appears in many fields of engineering when the goal is to recover ``hidden'' characteristics of a system from ``observed'' data. In recent years, deep learning techniques have shown remarkable success for solving inverse problems in various fields such as compressed sensing, medical imaging \cite{senouf2019self}, and many more (see~\cite{ongie2020deep} for a recent overview). In general, the inverse problem is ill-posed, i.e., one may not be able to uniquely recover the input field given noisy and incomplete observations~\cite{ongie2020deep,gilton2021deep}. Moreover, traditional methods for solving $z$ are compute intensive, as they require a large number of model runs (especially if $z$ has a large dimension). 

In this paper, we propose to compute $\boldsymbol{z}$ given $\boldsymbol{X^t}$ and $\boldsymbol{y^t}$ efficiently. Deep learning methods traditionally solve inverse problems by minimizing a cost function \cite{ma2020deep} that consists of a data-fit term, which measures how well the reconstruction matches the observations and a regularizer. These methods, largely based on convolution operator tend to work for linear inverse problems such as image denoising, super-resolution, and compressed sensing, but for capturing time varying physical processes such as ours (non-linear inverse problem), the traditional method fails. This motivates us to ask the question: \emph{Can we develop a robust inverse framework that can accurately recover static features when they are noisy/missing/uncertain through an inverse modeling process ?} 

To answer this question, we design a novel inverse framework leveraging knowledge from the hydrological domain in a self-supervised learning framework to implicitly extract complex correlations embedded in the input data. We call our methodology knowledge-guided self supervised learning (KGSSL). We show that KGSSL can be used to reduce the uncertainty in the measurement of basin characteristics, impute basin characteristics and identify unknown basin characteristics. KGSSL enables the extraction of time-invariant characteristics autonomously by distinguishing learning between the same basin to be similar for different years and dissimilar with other basins. In cases where certain basin characteristics are known, we further add a pseudo-inverse loss on top of the learned embeddings to guide the learning using the known basin characteristics. 

In the rest of our paper, we demonstrate the usefulness of KGSSL on real streamflow data (CAMELS~\cite{newman2015gridded}). Specifically, our method is used to identify existing and estimate new characteristics of the physical system. If the system's characteristics are missing or known but uncertain, then our methodology can impute them or reduce their uncertainty. We additionally show its usefulness when for any new basins, we don't have any knowledge of static characteristics $\boldsymbol{z}$. We show empirical results on the three research questions raised in this paper, namely measurement uncertainty, imputation, and unknown nature of the basin characteristics. We also present the proposed framework as a potential solution for performing source specific modulation in the driver-response model. Our main contributions are listed below:

\begin{itemize}
    \item We demonstrate the power of leveraging domain knowledge between entities in the context of an inverse problem.
    \item We perform extensive experiments on real streamflow data and show that  KGSSL outperforms baseline by 16 \% in reconstructing characteristics.
    \item We further show that our approach is highly robust to uncertainty in the basin characteristics. KGSSL outperforms the state-of-the-art forward model by 35\% when plugging KGSSL inferred characteristics. 
\end{itemize}

\section{Related Works}
Inverse problems \cite{ongie2020deep} always exist together with their forward problem. 
The goal of the inverse problem is to recover "hidden" information (which we cannot observe directly or is very expensive to observe) from readily available "observed" data.  Unfortunately, the inverse is often both intractable and ill-posed, since crucial information is lost in the forward process. However, the inverse process is required to inform us about physical parameters of the system (e.g., mass, temperature, physical dimensions, or structure), sources of influence, reconstruction of the coefficients in the equations that we cannot observe otherwise. Inverse problems are studied for many environmental science branches, i.e., hydrogeology \cite{zhou2014inverse}, geophysics  \cite{kim2018geophysical}, oceanography \cite{woolway2021winter}, meteorology \cite{pecha2021determination}, remote sensing \cite{dao2021improving}, etc. For example, an inverse problem arises when we reconstruct Earth's interior by modeling the physical propagation of seismic waves \cite{ritsema2000seismic}. Similarly, in reservoir engineering \cite{ganachaud2000improved}, given various measurements of geophysical fields, an inverse problem arises to determine the subsurface properties, such as the permeability field. Most of the recent deep learning approaches \cite{ardizzone2018analyzing,asim2020invertible} model forward/inverse mapping within a single network. However, in hydrology, the initial physical parameters are not known reliably \cite{kumarasamy2018calibration} for the basins/catchment due to temporal and spatial heterogeneity. This leads to a noisy forward operator, which makes existing inverse approaches ineffective. This motivates us to design a robust inverse framework impervious to corrupted basin characteristics. 

Due to abundant unlabeled data in computer vision, recently, researchers have started investigating self-supervised methods \cite{jing2020self} for model training. In self-supervised learning, the models are trained using pretext tasks instead of an actual task. For example, image colorization \cite{larsson2017colorization}, image inpainting \cite{pathak2016context}, solving image-jigsaw \cite{noroozi2016unsupervised}, learning by counting \cite{noroozi2017representation}, predicting rotations \cite{gidaris2018unsupervised}, etc. For a comprehensive understanding of self-supervised representation learning, we would like to redirect the reader to a survey by Jing et al.~\cite{jing2020self}. Our work utilizes a self-supervised loss called InfoNCE loss~\cite{chen2020simple} to counter the uncertainty in the basin characteristics by implicitly extracting complex correlations embedded in the meteorological drivers. The use of InfoNCE loss in our work is closely related to that of \cite{ravula2021inverse}, which trained the teacher-student network using contrastive loss to recover the true feature of a corrupted image. However, our problem domain (learning relationship between time-varying complex physical processes) is fundamentally different from vision-related inverse problems. In addition, our method differs in the following aspects. First, we employ the self-supervised learning method in the time-series domain, whereas most of the applications are in the vision domain. Second, we design novel pretext tasks in hydrology using domain knowledge. Finally, we focused extensively on robustness in our work and showcased our methodology's use on both supervised and unsupervised learning. 

\vspace{-.2in}
\section{Method}
In this work, we study the driver-response relation in dynamical systems. Specifically, we assume a dataset consisting of $N$ entities (an entity can be a lake, basin or streams in a river-network). For each entity $i$, the daily drivers are represented by $\boldsymbol{X_i}$ as a multivariate time series for $T$ timestamp i.e. $\boldsymbol{X_i} = [\boldsymbol{x^1_i}, \boldsymbol{x^2_i}, \dots, \boldsymbol{x^T_i}]$ where $\boldsymbol{x^t_i} \in \mathbb{R}^{D_x}$ indicates input vector at time $t \in T$ with $D_x$ dimension. $\boldsymbol{z_i} \in \mathbb{R}^{D_z}$ denotes the static characteristic vector of an entity with $D_z$ dimensions. The transient response corresponding to ${(\boldsymbol{X_i}, \boldsymbol{z_i})}$ for an entity is denoted by $\boldsymbol{Y_i} = [y^1_i, y^2_i, \dots, y^T_i]$.

Our proposed method KGSSL, infers time-invariant entity characteristics ($\boldsymbol{z_i}$) given the time-varying driver ($\boldsymbol{X_i}$) and response ($\boldsymbol{Y_i}$) data. KGSSL has several components. First, a \textit{Sequence Encoder} is used to extract a fixed length representation from the driver-response time-series. Second, a \textit{reconstruction loss} ($\mathcal{L}_{Rec}$) that forces the fixed length representation to capture the information stored in driver-response time-series by penalizing bad driver-response time-series reconstructions. Third, a \textit{Knowledge-guided Contrastive Loss} ($\mathcal{L}_{Cont}$) that implicitly extract complex correlations embedded in the driver-response time-series and enforces the physical knowledge that the entity characteristics are time-invariant. Finally, a \textit{PseudoInverse loss} ($\mathcal{L}_{Inv}$) that encourages robust reconstruction of entity characteristics from the fixed length representation using a feed-forward network. Thus, the final loss function for training KGSSL is 
\begin{equation}
    \label{eqn:representation learning}
    \mathcal{L} = \lambda_1 \mathcal{L}_{Rec} + \lambda_2 \mathcal{L}_{Cont} + \lambda_3 \mathcal{L}_{Inv}
\end{equation}
where $\lambda_1$, $\lambda_2$, $\lambda_3$ are hyper-parameters to control the weights of three loss terms. $\mathcal{L}_{Inv}$ is added only when the entity characteristics are known and available for training. We can also train KGSSL using only the self-supervised loss functions, $\mathcal{L}_{Rec}$ and $\mathcal{L}_{Cont}$. In the following subsections, we discuss each of these components in detail and provide intuition behind such design choices.

KGSSL generates time invariant and entity specific embeddings from driver-response time-series data. Specifically, for each entity $i$ in a given set of N entities, we randomly select two sequences of length $W$. Let $S_{a_i}$ and $S_{p_i}$ be the two sequences taken from the time-windows $t_{a_i}:t_{a_i}+W$ and $t_{p_i}:t_{p_i}+W$, respectively. This results in $2N$ sequences and each element in these sequences are formed by concatenating the drivers-response time-series of the entity ($[\boldsymbol{x^t_i};y^t_i]$).

\begin{figure}[t!]
    \centering
    \includegraphics[width=0.7\linewidth]{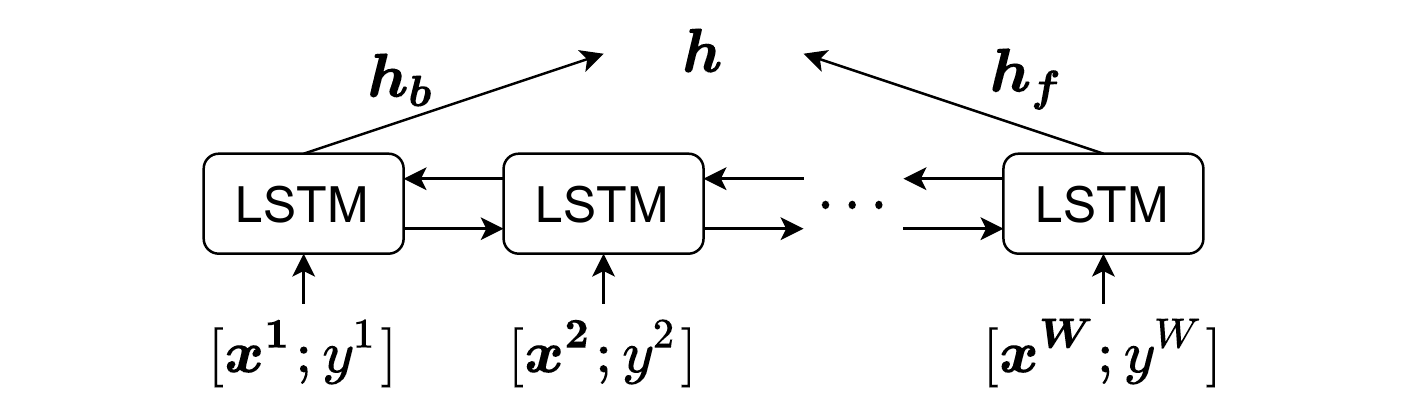}
    \caption{Bidirectional LSTM based Sequence Encoder}
    \label{fig:LstmEncoder}
\end{figure}

\subsection{Sequence Encoder}
We use a sequence encoder to encode the temporal information and the interaction between the driver and response in these sequences. LSTM is particularly suited for our task where long range temporal dependencies between driver and response exist as they are designed to avoid exploding and vanishing gradient problems. However, LSTMs are designed to run forward in time and cannot provide explainability on the current time-steps given the future data. To capture this information we use a Bidirectional LSTM based sequence encoder $\mathscr{E}$ (Figure~\ref{fig:LstmEncoder}). Specifically, we build two LSTM structures:the forward LSTM and the backward LSTM. The two LSTM structures are the same except that the time-series is reversed for the backward LSTM. Each LSTM uses the following set of equations to generate the embeddings for a sequence,

\begin{equation}
    \begin{split}
        \boldsymbol{i_t}    &= \sigma (\boldsymbol{W_i}\left[[\boldsymbol{x^t};y^t];\boldsymbol{h^{t-1}}\right] + \boldsymbol{b_i})\\
        \boldsymbol{f_t}    &= \sigma (\boldsymbol{W_f}\left[[\boldsymbol{x^t};y^t];\boldsymbol{h^{t-1}}\right] + \boldsymbol{b_f})\\
        \boldsymbol{g_t}    &= \sigma (\boldsymbol{W_g}\left[[\boldsymbol{x^t};y^t];\boldsymbol{h^{t-1}}\right] + \boldsymbol{b_g})\\
        \boldsymbol{o_t}    &= \sigma (\boldsymbol{W_o}\left[[\boldsymbol{x^t};y^t];\boldsymbol{h^{t-1}}\right] + \boldsymbol{b_o})\\
        \boldsymbol{c_t}    &= \boldsymbol{f_t} \odot \boldsymbol{c_{t-1}} + \boldsymbol{i} \odot \boldsymbol{g_t}\\
        \boldsymbol{h_t}    &= \boldsymbol{o_t} \odot \tanh{(\boldsymbol{c_t})}\\
    \end{split}
\end{equation}

Each of the forward and backward LSTM takes in a sequence $S$ as input and generates corresponding embeddings $\boldsymbol{h_f}$ and $\boldsymbol{h_b}$ $(\boldsymbol{h}=\mathscr{E}(S))$. These embeddings are essentially the final hidden states of each LSTM. The embeddings for the forward LSTM ($\boldsymbol{h_f}$) and backward LSTM ($\boldsymbol{h_b}$) are added to get the final embeddings $\boldsymbol{h}$ as shown in Figure~\ref{fig:LstmEncoder}. These embeddings capture the temporal information as well as the driver-response interaction by modeling the change in streamflow due to the weather drivers in both forward and backward directions.

\subsection{Reconstruction Loss}
To preserve the key information from driver-response data, we use a standard LSTM based decoder $\mathscr{D}$ that reconstructs the sequence back from the embedding ($\hat{S} = \mathscr{D}(\boldsymbol{h})$). The LSTM decoder uses its own output at the previous time-step as the input for the current time-step and thus can be regarded as a sequence generator using the embedding $\boldsymbol{h}$ as a prior. The reconstruction error is computed as the mean-squared error between the reconstructed and the original sequence, as shown below,
\begin{equation}
    \mathcal{L}_{Rec} = \frac{1}{2N} \sum_{e\in \{a,p\}} \sum_{i=1}^N MSE(\hat{S}_{e_i}, S_{e_i})
\end{equation}

Here $\mathcal{L}_{Rec}$ acts as a regularizer in representation learning, by extracting meaningful information from the time-varying input data. However, since we are interested in extracting the time-invariant information from the time-series data, solely relying on $\mathcal{L}_{Rec}$ leads to sub-optimal performance. $\mathcal{L}_{Rec}$ promotes preservation of information about the time-series in the embeddings which will later be used by the decoder to reconstruct back the input time-series.

\subsection{Knowledge-guided Contrastive Loss}
Each entity's response to a given driver is governed by complex physical processes captured by its inherent physical characteristics that remain constant through time. Moreover, different entities have different responses to the same driver due to the differences in their inherent characteristics. We use this physical knowledge  of entities to define a self-supervised contrastive loss~\cite{chen2020simple,oord2018representation}. Specifically, the sequences $S_{a_i}$ and $S_{p_i}$ of an entity form a positive pair, and for each positive pair, we treat the other 2(N-1) sequences within a batch as negative examples. Thus, the contrastive loss forces the embeddings $\boldsymbol{h_{a_i}}$ and $\boldsymbol{h_{p_i}}$ resulting from the sequences $S_{a_i}$ and $S_{p_i}$ of the same entity to be similar and different from the embeddings of other basins. For a given positive pair, the loss is calculated as,
\begin{equation}
    \begin{split}
        l(a_i,p_i) = & \frac{\exp{(sim(\boldsymbol{h_{a_i}}, \boldsymbol{h_{p_i}})/\tau)}}{\sum_{e\in\{a,p\}}\sum_{j=1}^N\exp{(sim(\boldsymbol{h_{a_i}}, \boldsymbol{h_{e_j}})/\tau)}}\\
        + &\frac{\exp{(sim(\boldsymbol{h_{p_i}}, \boldsymbol{h_{a_i}})/\tau)}}{\sum_{e\in\{a,p\}}\sum_{j=1}^N\exp{(sim(\boldsymbol{h_{p_i}}, \boldsymbol{h_{e_j}})/\tau)}}
    \end{split}
\end{equation}
where, $sim(\boldsymbol{h_{a_i}}, \boldsymbol{h_{p_i}})=\frac{\boldsymbol{h_{a_i}}^T\boldsymbol{h_{p_i}}}{\|\boldsymbol{h_{a_i}}\|\|\boldsymbol{h_{p_i}}\|}$. Thus, the total contrastive loss for 2N such positive pairs is given as,
\begin{equation}
    \mathcal{L}_{Cont} = \frac{1}{2N} \sum_{i=1}^N l(a_i,p_i)
\end{equation}

Both $\mathcal{L}_{Cont}$ and $\mathcal{L}_{Rec}$ do not require any supervised information and thus can work with a large number of entities for which we only know the driver-response time-series. Moreover, later in the results, we show that using only one of these losses leads to sub-optimal performance and, thus we use a combination of these two losses.

\begin{figure}[t!]
    \centering
    \includegraphics[width=\linewidth]{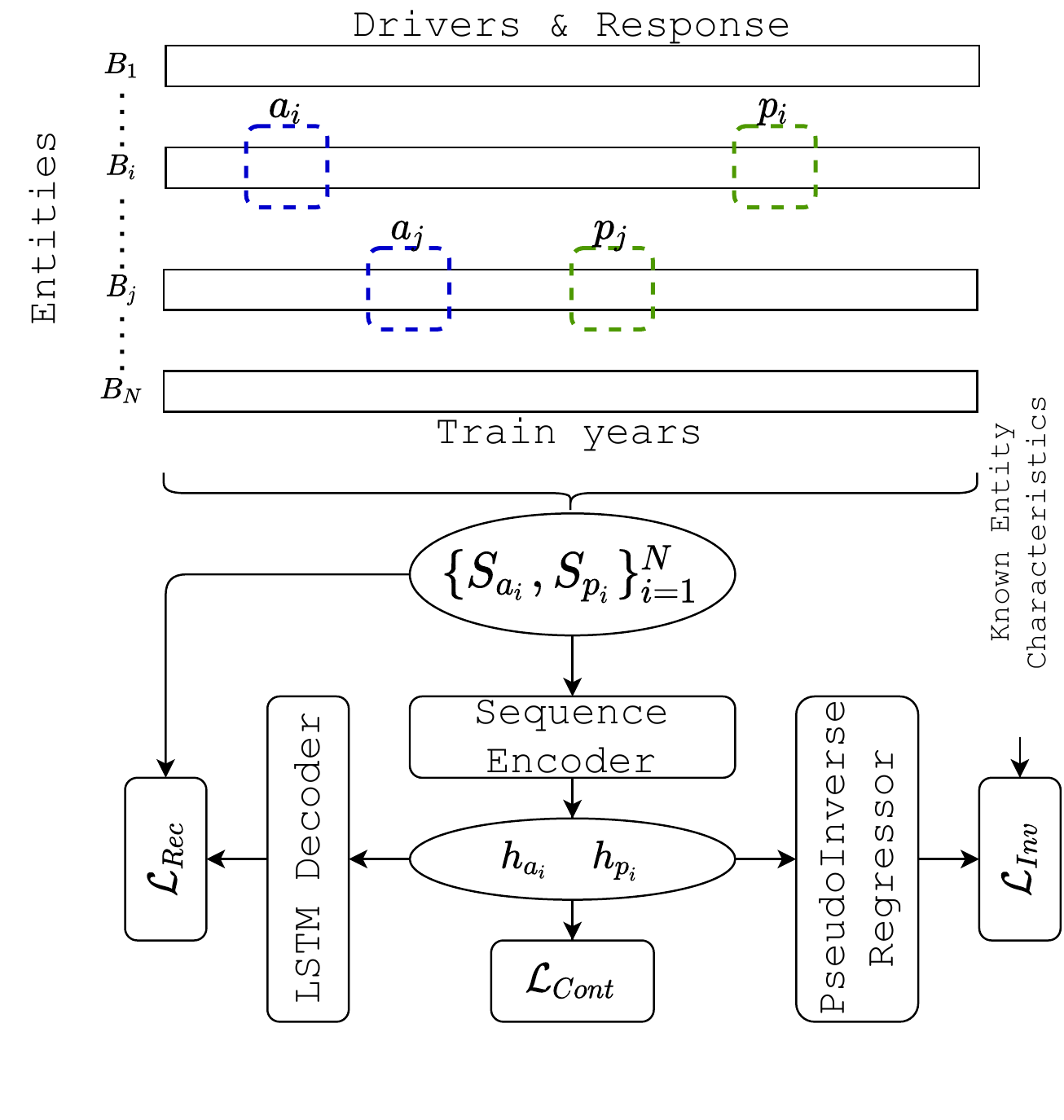}
    \vspace{-.4in}
    \caption{\small The proposed inverse model generates embeddings for a basin from the LSTM Encoder (Figure~\ref{fig:LstmEncoder})and is trained in a self-supervised manner. Strong supervision $(\mathcal{L}_{Inv})$ is added when ground-truth characteristics are available for a limited number of basins}
    \label{fig:Architecture}
    \vspace{-.2in}
\end{figure}

\subsection{PseudoInverse Loss}
Reconstruction Loss and Knowledge-guided Contrastive Loss is used to extrapolate entity characteristics from the time-varying driver ($\boldsymbol{X_i}$) and response ($\boldsymbol{Y_i}$). However, if some entity characteristics are known (albeit noisy~\cite{song2020learning}/uncertain), the above loss functions fail to account for them during the model training. To improve our inverse framework, we propose using PseudoInverse loss that utilizes incomplete/uncertain missing entity characteristics as a source of supervision. Specifically, we add a feed-forward layer $\mathscr{I}$ on sequence encoder output $h$ to estimate $\boldsymbol{\hat{z}} = \mathscr{I}(\boldsymbol{h})$ and then we define regression loss with the available set of entity characteristics ($z$) as shown in Figure~\ref{fig:Architecture} . Pseudoinverse loss is defined as follows: 
\begin{equation}
    \mathcal{L}_{Inv} = \frac{1}{N} \sum_{i=1}^N \frac{1}{z} \sum_{j=1}^z (z_i^j-\hat{z}_i^j)^2
\end{equation}

\subsection{Reconstructing static characteristics given temporal data}\label{recon_static}
Our KGSSL framework can be used to generate entity-specific embeddings as well as static characteristics. Specifically, given input-drivers $X_i = [\boldsymbol{x^1_i}, \boldsymbol{x^2_i}, \dots, \boldsymbol{x^T_i}]$ where $\boldsymbol{x^t_i} \in \mathbb{R}^{D_x}$ and output-response $Y_i = [y^1_i, y^2_i, \dots, y^T_i]$ time-series of length $T$ for an entity, we break the combined time-series into $T/W$ sequences of length $W$. Each of these sequences $S_i^j$ are fed to the encoder $\mathscr{E}$ to generate an embedding $\boldsymbol{h}_i^j$, which are further fed into the inverse regressor $\mathscr{I}$ to predict the static characteristics $\boldsymbol{\hat{z}}_i^j$. By taking the element-wise mean of the embeddings, we get the final embeddings of the entity. Similarly, we get the final estimate of the static characteristics along with their uncertainties by taking the element-wise mean and standard deviation of the sequence specific predictions, as shown,
\begin{equation}
    \label{eqn:unc}
    \boldsymbol{h}_i = \frac{W}{T}\sum_{j=1}^{T/W} \boldsymbol{h}_i^j \qquad \boldsymbol{\hat{z}}_i = \frac{W}{T}\sum_{j=1}^{T/W} \boldsymbol{\hat{z}}_i^j \qquad \boldsymbol{unc}_i = \sqrt{\frac{W}{T}\sum_{j=1}^{T/W} (\boldsymbol{\hat{z}}_i^j-\boldsymbol{\hat{z}}_i)^2}
\end{equation}

As more and more years of data are made available for an entity, the embeddings and the predictions of the static characteristics become more certain and informative.


\begin{figure}[t]
    \centering
    \includegraphics[width=0.6\linewidth]{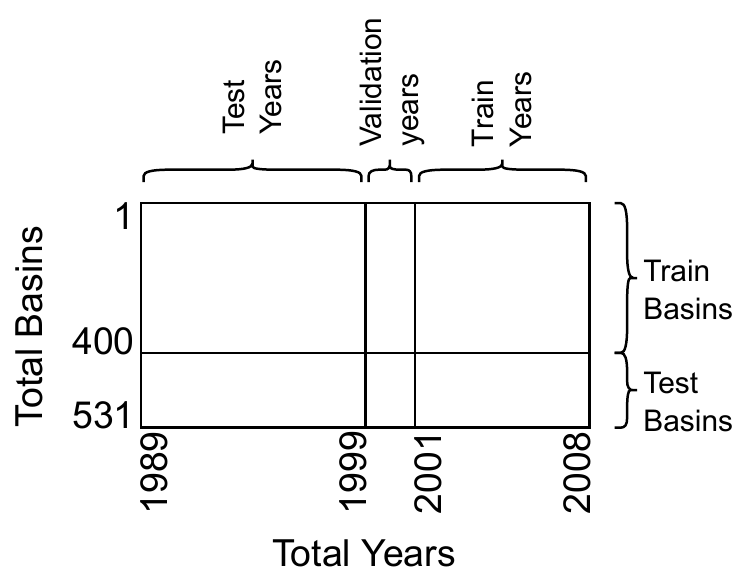}
    \vspace{-.2in}
    \caption{\small Experimental setting followed in the paper for training and testing of the ML models}
    \label{fig:experimentalSetting}
    \vspace{-.2in}
\end{figure}

\section{Experimental Results}
\subsection{Datasets and Implementation details}
We evaluate KGSSL using the CAMELS (Catchment Attributes and MEteorology for Large-sample Studies) dataset, which is extensively used for investigating hydrology processes, in particular, streamflow prediction~\cite{Addor2017}. CAMELS compiles meteorological forcing data (e.g. precipitation, air temperature), streamflow observation, calibrated physical model simulation, and catchment characteristics(see Appendix A.1 for a complete list), all of which makes it possible to leverage recent developments in machine learning, in particular deep learning, in the hydrology community to advance continental hydrology modeling~\cite{kratzert2019towards,feng2020enhancing}. In particular, using the CAMELS dataset, Kratzert et al. \cite{kratzert2019towards} showed that a global scale LSTM model (that uses known static characteristics as input in addition to weather drivers) can outperform state-of-the-art physics based hydrological model that are individually caliberated for each basins.

Following the set up used by Kratzert et al. ~\cite{kratzert2019towards}, our study uses data for 531 basins from CAMELS for the periods (1989-2009). Of these, (2001-2008) is used for model building, and the rest is used for testing. Fig.~\ref{fig:experimentalSetting} show the experimental setting followed in this paper. Like Kratzert et al. our study uses 27 basin characteristics organized by physically meaningful groups: climatology, soils/geologic conditions and geomorphology/land-cover. These three groups of characteristics can generally be assumed to represent physical characteristics that contribute more or less to the rainfall/runoff process in any given catchment.

\begin{figure*}[t]
    \centering
    \includegraphics[width=\linewidth]{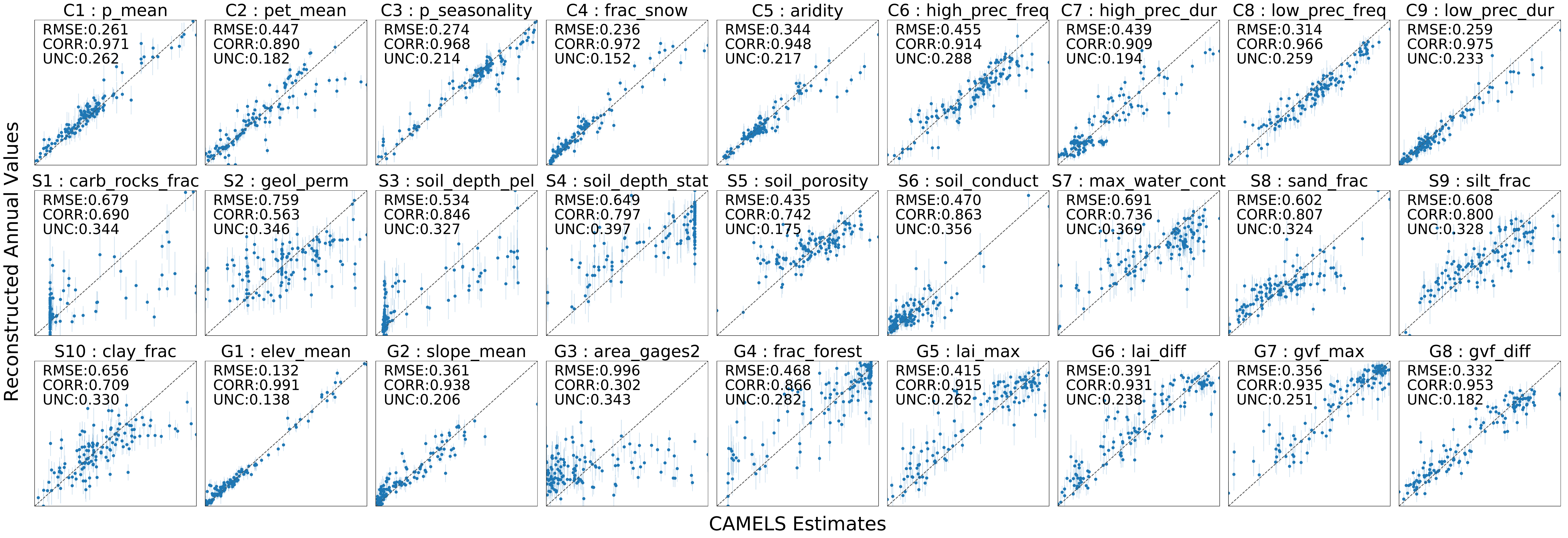}
    \vspace{-0.3in}
    \caption{\small Scatter plot of the CAMELS Estimates for static characteristics (x-axis) for the testing basins vs. reconstructed annual characteristics (y-axis). The error bar across the points show the variation of the reconstructed characteristics annually across test years.}
    \label{fig:PredictChars}
\end{figure*}

\begin{table}[]
    \centering
    \begin{tabular}{|c|c|c|}
        \hline
        \textbf{Method}     & \textbf{RMSE}     & \textbf{CORR}\\
        \hline
        LSTM                & 0.540             & 0.795 \\
        \hline
        KGSSL($L_{Rec+Inv}$)    & 0.493             & $\mathbf{0.831}$\\
        \hline
        KGSSL($L_{Cont+Inv}$)    & 0.514             & 0.815\\
        \hline
        KGSSL($L_{Rec+Cont+Inv}$)  & $\mathbf{0.465}$  & 0.824\\
        \hline
    \end{tabular}
    \caption{\small Average root mean square error (RMSE) and correlation (CORR) for $131$ test basins during testing period.} 
    \label{tab:estimateentity}
    \vspace{-.3in}
\end{table}

We create input sequences of length 365 using a stride of half the sequence length, i.e., 183. This results in 13 windows for the data used for model training and 19 for the testing period. All LSTMs used in our architecture have one hidden layer with 64 units. The feed-forward network to reconstruct characteristics has one hidden layer followed by activation to introduce non-linear combinations of the embeddings. The hyperparameter $\lambda_1$, $\lambda_2$, and $\lambda_3$ are set at 1, 1, and 1 respectively. The value of $\lambda_1$, $\lambda_2$ and $\lambda_3$ are selected to balance the supervised and unsupervised components of the loss function. Higher values for $\lambda_3$ lead to lower training loss but at the expense of loss of robustness to noise in the static characteristics(see Appendix A.2 for more details about the hyperparameter search). To reduce the randomness typically expected with network initialization, we train five models with different initialization of deep learning model weights. The predictions were then further combined into an ensemble by averaging prediction from these five models.

In sections 4.2,4.3,4.4 we evaluate the ability of KGSSL to estimate the entity characteristics in test basins under various conditions, including when characteristics in the training set are corrupted or missing. Section 4.4 considers the case where the catchment characteristics are not available during training. In addition, section 4.6 shows the ability of KGSSL to improve the forward modeling task. 

\subsection{Estimating the entity characteristics}
\label{sec:estimate}
We train our model using 400 train basins and reconstruct the static characteristics of the remaining 131 test basins. Table \ref{tab:estimateentity} reports average root mean square error (RMSE) and correlation (CORR) for $131$ test basins during testing period.The entities have different scale values. Since the RMSE value is not scale-invariant, we also report a correlation metric, which is scale-independent. Moreover, the RMSE value measures prediction error, whereas correlation captures the trend.

\begin{table*}[t]
    \small
    \centering
    \begin{tabular}{|c|c|c|c|c|c|c|c|}
        \hline
        \multirow{2}{*}{Group}  & \multirow{2}{*}{Indexes}  & Sec \ref{sec:estimate}    & \multicolumn{3}{c|}{Sec \ref{sec:corrupted}}                                                              & \multicolumn{2}{c|}{Sec \ref{sec:missing}}\\
        \cline{3-8}
                                    &                       & Original                  & 90\% $\mathcal{N}(0,\sigma_i)$    & 50\% $\mathcal{N}(0,2\sigma_i)$   &90\% $\mathcal{N}(0,2\sigma_i)$   & 50\% Missing  & 90\% Missing\\
        \hline
        Climate                     &$C1-C9$                & 0.935                     & 0.906                             & 0.890                             & 0.854                             & 0.933         & 0.870\\
        Soil-Geology                &$S1-S10$               & 0.711	                    & 0.658	                            &0.585                              & 0.546                             & 0.665	        & 0.550\\
        Geomorphology-land cover   	&$G1-G8$        	    & 0.841	                    & 0.812	                            &0.783                              & 0.751                             & 0.825	        & 0.783\\
        \hline
        \multicolumn{2}{|c|}{Mean}                          & 0.824	                    & 0.786	                            &0.745                              & 0.709                             & 0.802	        & 0.725\\
        \hline
    \end{tabular}
    \caption{\small The correlation of reconstructed characteristics for test basins in reference to true characteristics for different levels of noise and missing values in train basins.}
    \label{tab:robustness}
    \vspace{-0.2in}
\end{table*}

We make the following high-level observations from our results: a) KGSSL, which uses both supervised and unsupervised loss functions to infer entity characteristics, has superior performance ($\mathbf{16\%}$ better RMSE) as compared to LSTM, which was trained using mean square error loss. b)Each of the self-supervised losses, i.e., $L_1$ and $L_2$ individually, leads to sub-optimal performance; thus, combining these two losses with $L_3$ helps capture the complex physical process accurately. Fig.~\ref{fig:PredictChars} illustrates the ability of KGSSL to reconstruct the 27 basin characteristics in the CAMELS dataset. Note that the reconstructed values are annual averages for each year in the testing period 1989-1999, while vertical lines show the uncertainty (UNC) in the prediction as described in Eq.~\ref{eqn:unc}). Each individual scatter plot showcases RMSE, Correlation (CORR) and uncertainty (UNC). Note that in general KGSSL performs well (correlation>0.8) in 20 out of 27 cases with correlation > 0.9 for 14 of them, and of the remaining 7 cases only one has a correlation <0.5.

The results (Table \ref{tab:estimateentity} and Figure \ref{fig:PredictChars}) exhibit that KGSSL is able to predict the characteristics with acceptable accuracy (corr>0.8) for most of the characteristics. In general, the average RMSE and average correlation of the predicted values are 0.465 and 0.824, respectively. However, the characteristics reconstruction performance varies among the individual characteristics. The ones with higher reconstruction RMSEs are usually accompanied with higher standard deviation. The features with satisfactory reconstruction performance (lower RMSEs and high correlation) are also more temporally consistent (lower standard deviation). As discussed in the next paragraph, inconsistencies in reconstruction performance among the individual characteristics can be reasoned based on domain knowledge and reflect uncertainties present in the original CAMELS data set. This interpretation of the modeling results is arguably an important scientific discovery of our proposed KGSSL framework.

KGSSL inferred all nine climate characteristics quite accurately. This result is consistent with the fact that the climate characteristics published in the CAMELS data set are derived directly from the meteorological forcing data $\boldsymbol{X_i}$. We reason that the \textit{elev\_mean} was also quite accurately inferred (0.132 RMSE and 0.138 standard deviation) because the mean elevation is related to climate patterns, and this reasoning also holds true for the catchment slope characteristic, \textit{slope\_mean}, and vegetation characteristics (\textit{gvf\_diff, gvf\_max, lai\_max, and lai\_diff, frac\_forest}), as these should all be correlated to meteorological characteristics. The remaining seven characteristics are uncertain by nature because of involved spatial and temporal heterogeneities. Some of them also possess uncertainties in the original data source from which they are derived. Most of these remaining characteristics are soil-related (e.g., \textit{carbonate\_rocks\_frac}, \textit{geol\_permeability}, \textit{soil\_depth\_pelletier}) and are derived as spatial averages from the catchments. Such derivation overly simplifies catchment spatial heterogeneity, in particular for large catchments. Therefore, this simplification might explain the large variance recognized in those characteristics. Furthermore, as mentioned in \cite{Addor2017}, the spatial gridded data where those soil characteristics are derived are uncertain and erroneous in certain geographic regions. It also only characterizes top layer soils and ignores deep soil information. Consequently, soil related characteristics are poorly constructed ones. In addition, \textit{area\_gages2} is the contributing area where surface runoff is generated, and this is spatially and temporally highly non-uniform due to the spatial variability of soil properties, spatial variability of antecedent conditions, and non-uniformity of incident rainfall. Thus, our reconstruction performance on \textit{area\_gages2} is also unsatisfactory. 

\begin{figure}
    \centering
    \includegraphics[width=\linewidth]{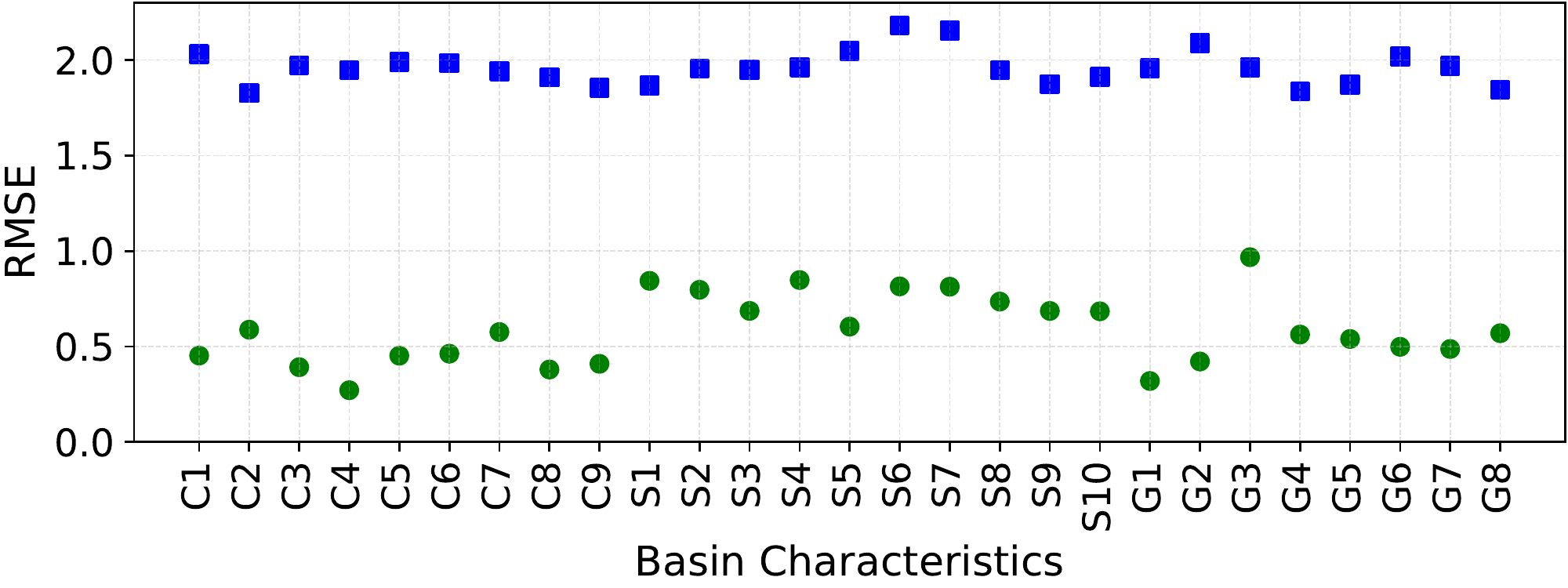}
    \caption{\small Comparison of the RMSE of the corrupted values (in blue) generated by adding Gaussian noise ($\mathcal{N}(0,2\sigma_i)$) to 50\% of basins and the reconstructed values (in green).}
    \label{fig:corr50}
    \vspace{-0.3in}
\end{figure}

\begin{figure*}[t]
    \begin{subfigure}{0.29\linewidth}
        \centering
        \includegraphics[width=\linewidth]{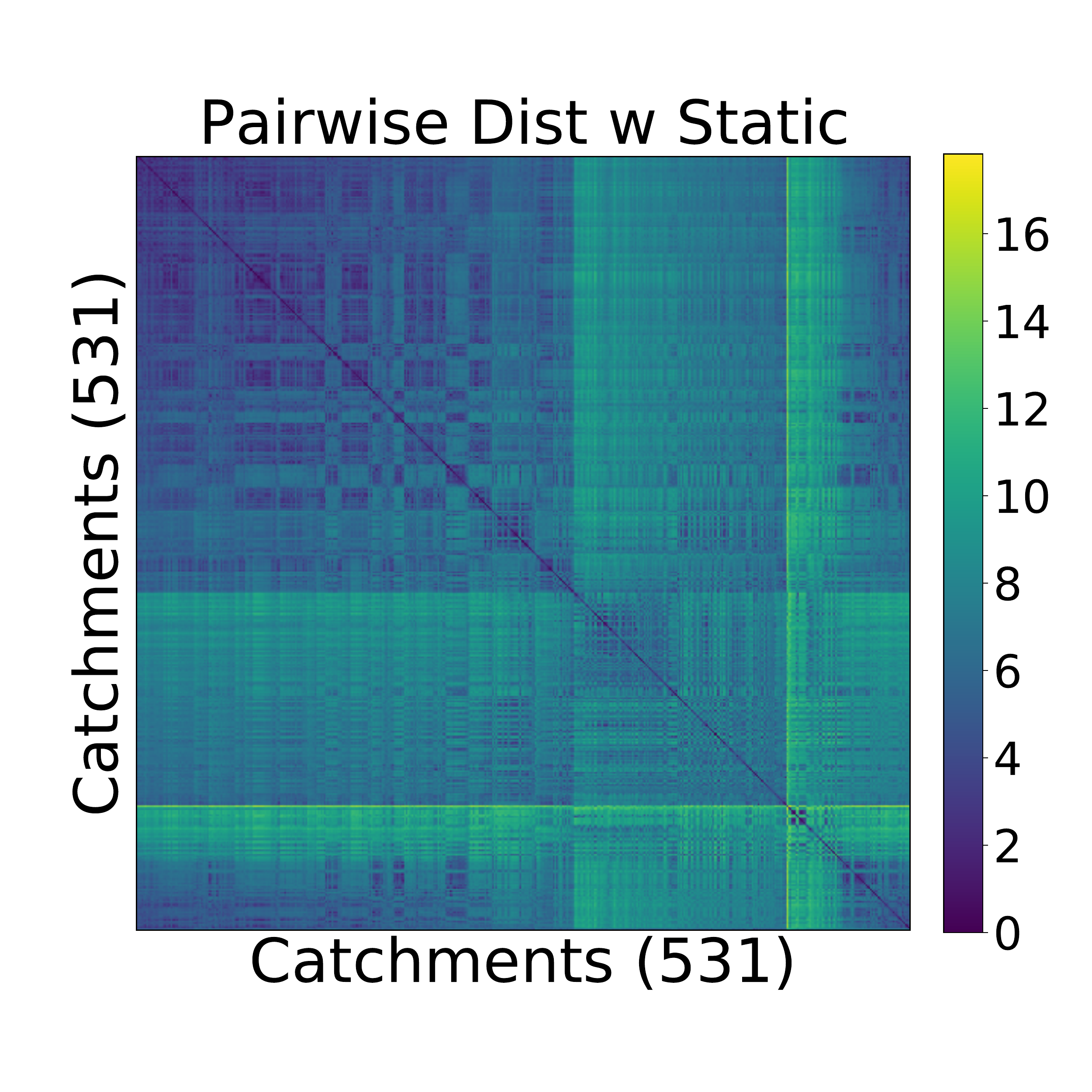}
        \vspace{-0.3in}
        \caption{}
        \label{fig:distanceStatic}
    \end{subfigure}%
    \begin{subfigure}{0.29\linewidth}
        \centering
        \includegraphics[width=\linewidth]{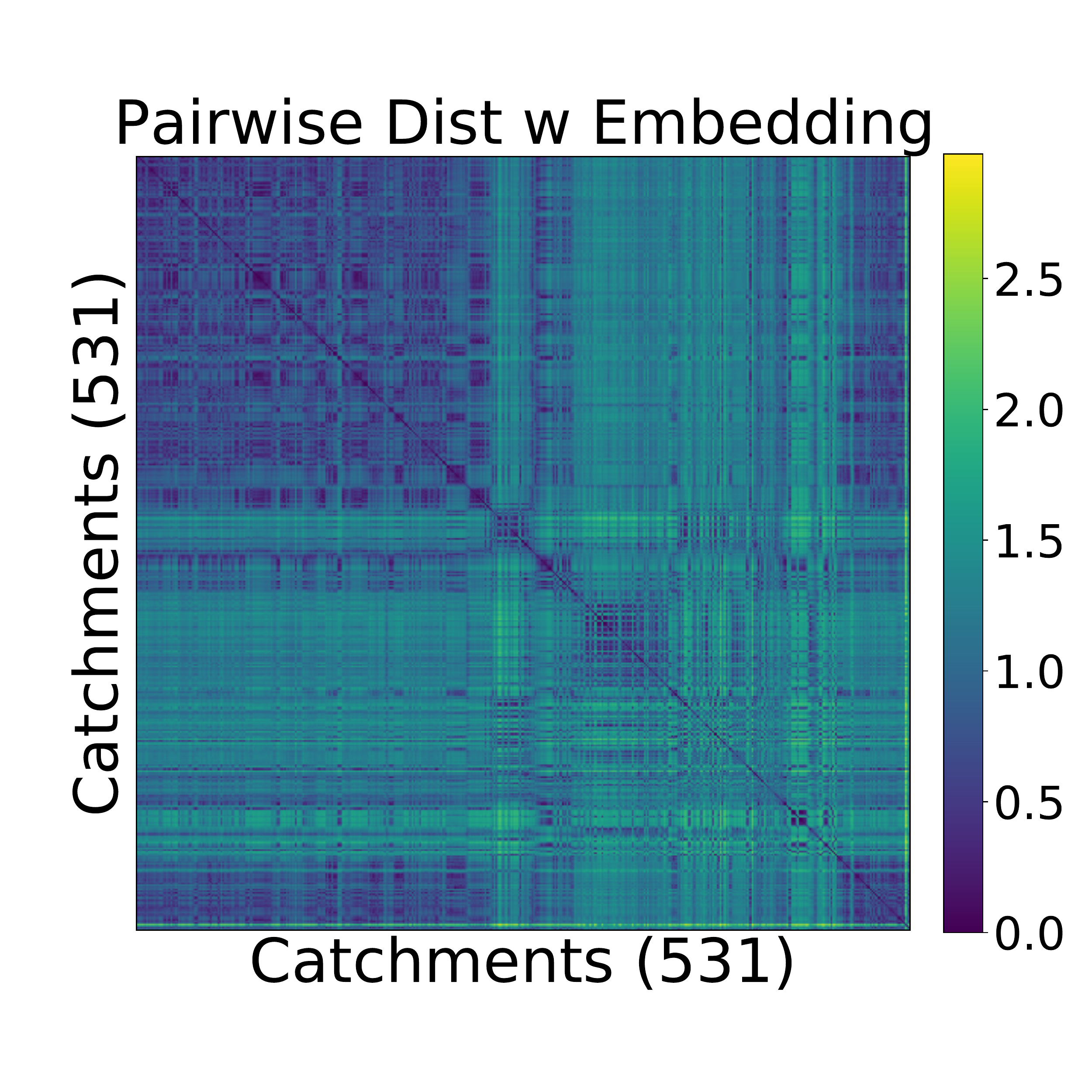}
        \vspace{-0.3in}
        \caption{}
        \label{fig:distanceEmbedding}
    \end{subfigure}%
    \begin{subfigure}{0.42\linewidth}
        \centering
        \includegraphics[width=\linewidth]{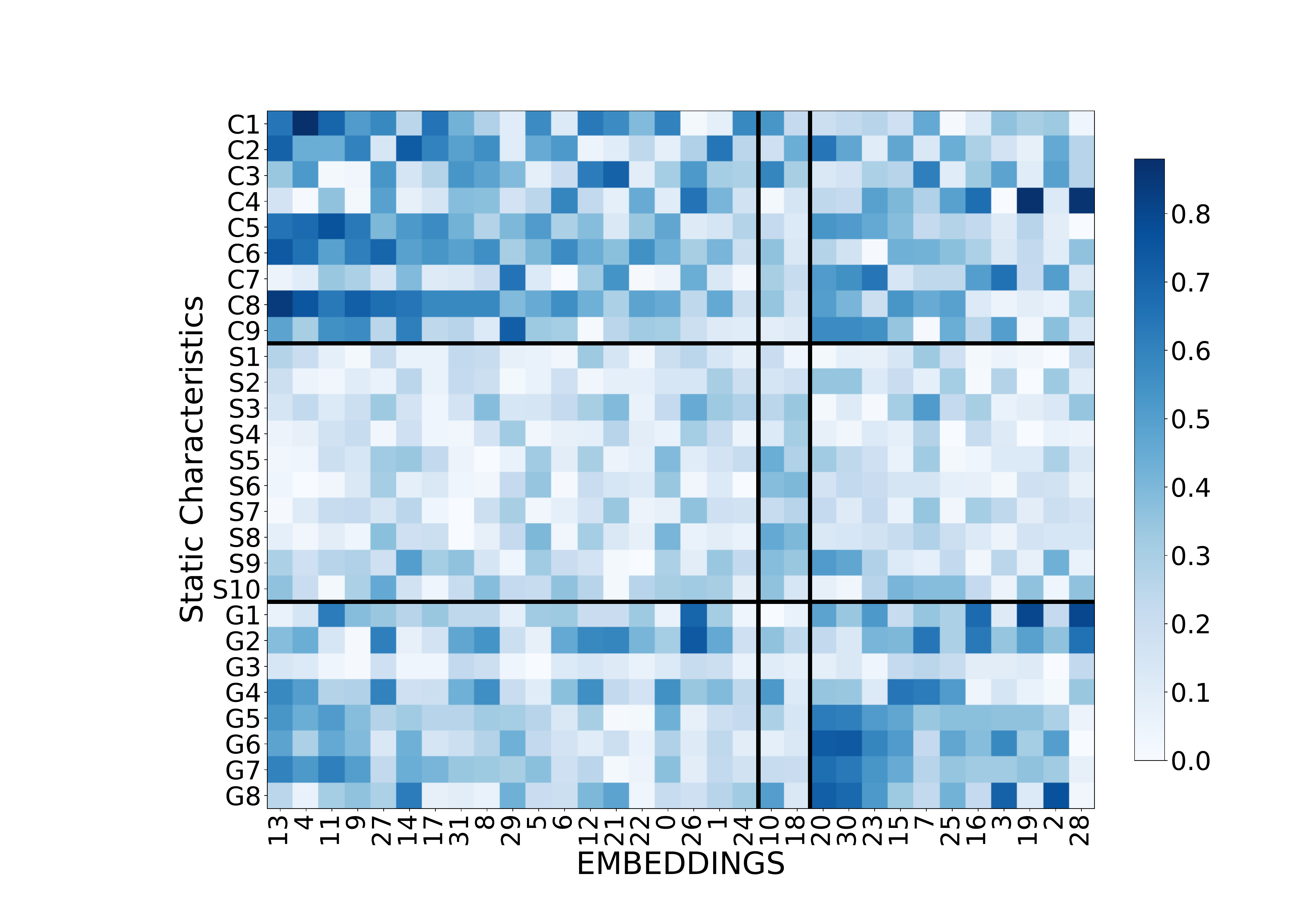}
        \vspace{-0.3in}
        \caption{}
        \label{fig:correlationStaticEmbedding}
    \end{subfigure}
    \vspace{-0.2in}
    \caption{\small Represent pairwise distance matrices for 531 catchments. Fig (a) Entry (i,j) is the pairwise distance between characteristic vector of catchments i and catchment j; Fig (b) Entry (i,j) is the pairwise distance between embedding vectors generated using KGSSL for catchment i and catchment j (c) Correlation of each dimension of the learned embeddings with each physical characteristic.}
\end{figure*}

\subsection{Robustness to Corruption in available characteristics}
\label{sec:corrupted}
As highlighted in Sec.\ref{Sec:Introduction}, we expect uncertainty in the characteristics, , and furthermore the nature of this uncertainty may be due to temporal and/or spatial variability, lack of representativeness, measurement error and/or missing data. The inverse model learns generalizable patterns and hence can potentially denoise the corrupted characteristics. To emulate this uncertainty in measurement we randomly corrupt 50\% and 90\% of the characteristics. Three experimental setting are thus created. First, to 50\% of the characteristics, a Gaussian noise with 0 mean and 2 standard-deviation is added while the remaining characteristics are left unchanged. Second, to 90\% of the characteristics a Gaussian noise with 0 mean and 1 standard-deviation is added. Finally, to the same 90\% of the characteristics a Gaussian noise with 2 standard-deviation is added. Those scenarios are created to capture two perspectives: a small number of characteristics can have a high level of noise, and a large number of characteristics are corrupted with a relatively low level noise. We train separate models on the training data using the corrupted values of these three settings, and the basin characteristics were predicted using the data from the test years for all the basins and compared to the original values.

We compare the performance of KGSSL trained using the corrupted data with the KGSSL trained using original catchment characteristics. Table~\ref{tab:robustness} shows the performance of various methods in terms of correlation of the predicted characteristics with the original characteristics. The impact of noise on characteristics varies among groups, which can be explained by their dependence on weather data that characteristics are learned. Climate characteristics are the least sensitive ones because their original characteristics are derived from weather data. Noisy original characteristics will not downgrade the reconstruction performance because characteristics can be learned from weather data anyway. Though not directly related to weather data, geomorphology-land cover characteristics exhibit geomorphology, land cover patterns that are implicitly characterized from weather data because of involved plant growing mechanisms and terrestrial processes. Thus, their reconstruction performance is much less impacted by noise in the training set. The worst responses in soil-geology characteristics are likely because they characterize subsurface processes whose interactions with weather data are relatively negligible. Such limited usable information in weather data for soil-geology characteristics constrains the capability of our model to learn. Figure~\ref{fig:corr50} shows the RMSE computed for corrupted (in blue) and reconstructed characteristics (in green) with respect to true characteristics averaged across all the 400 train basins for these two models. We can observe that KGSSL significantly reduces measurement error in characteristics by an average RMSE of 1.369.

\vspace{-0.1in}
\subsection{Robustness to Missing characteristics}
\label{sec:missing}
Representing physical processes, catchment characteristics often serve as a unique catchment signature. However, owing to the availability of various data sources, characteristics that represent one region are likely not available in another region. It therefore creates a common and important application scenario where a complete set of catchment characteristics across catchments are not assured. This limitation is more pronounced for cross-continental catchments whose characteristics are overlapping rather than exactly matching with each other. For instance, over half of catchment characteristics (e.g., main stream length, bulk density) in CAMELS-CN (a version of CAMELS for China)~\cite{Hao2021} are not included in the characteristics set of the catchments in the CAMELS dataset being used in this paper (which only contains basins from USA)~\cite{Addor2017}. The same scenario is also present in CAMELS version for Great Britain~\cite{Coxon2020}, Chili~\cite{Alvarez-Garreton2018}, and Brazil~\cite{Chagas2020}. In addition, insufficient understanding of catchment processes will also lead to a select set of characteristics that miss the opportunity to capture certain hydrological processes beyond current hydrological understanding. To address this issue, the KGSSL can potentially estimate catchment characteristics when they are missing for some catchments. To emulate such a scenario of missing catchment characteristics, we use a similar set up as in Sec.~\ref{sec:corrupted}. Instead of adding Gaussian noise, we treated 50\% and 90\% of the characteristics to be missing. We trained separate models for each of these settings on train years and train basins, where $\mathcal{L}_{Inv}$ was calculated and used for training the model only when characteristics were available. The catchment characteristics were predicted using the data from the test years for all the basins and compared to the original catchment characteristics.

The predicted catchment characteristics using data from the test years are compared to the original catchment characteristics. For the setting with 50\% missing data, the average RMSE and average correlation of the predicted values are 0.540 and 0.802 respectively, whereas for 90\% missing data, the average RMSE and average correlation of the predicted values are 0.646 and 0.725 respectively. This result suggests that KGSSL can potentially be used to impute the missing characteristics. Further, Table~\ref{tab:robustness} shows the robustness of our method, where we predict the characteristics for the 131 test catchments using the data from the test years using both the models and compare the prediction performance to the model trained using the clean data (Section~\ref{sec:estimate}).

\vspace{-0.1in}
\subsection{Discovering characteristics in the absence of ground truth(known characteristics)}
\label{sec:unknown}
Here we investigate the ability of KGSSL to identity time invariant characteristics that may be missing from available characteristics. We train the inverse model without using any knowledge of available characteristics that we can use as a constraint. $\mathcal{L}_{Rec}$ and $\mathcal{L}_{Cont}$ are used for training the model using the data from train years for all 531 basins. Further, using the data from the test years the embeddings for each basin are computed. To empirically demonstrate the characteristics captured by the learned embeddings, we calculate the pairwise-euclidean distance between two basins using their 27d physical characteristics (Figure~\ref{fig:distanceStatic}) and compare them with the distances computed using learned embeddings (Figure~\ref{fig:distanceEmbedding}). We generate (Figure~\ref{fig:distanceStatic}) by reordering the rows in the distance matrix computed using 27d physical characteristics such that basins with the least distances between themselves are placed close to each other to form a band-like structure. The exact order of basins used to generate (Figure~\ref{fig:distanceStatic}) is then further applied to the distance matrix computed using learned embeddings to generate (Figure~\ref{fig:distanceEmbedding}). From the figure we observe similar patterns in both the distance matrices which shows that KGSSL generates embeddings that contains meaningful similarity structure between basins. Further, we calculate the correlation between the learned embeddings with each of the physical characteristics for 531 basins. Figure~\ref{fig:correlationStaticEmbedding} provides a measure of the relative contribution of the 3 groups (C1-9 Climate, S1-10 Soils/ Geology, G1-8 Geomorphology/Landcover) for explaining the rainfall-runoff process. The vertical axis represents the 27 Static Characteristics, and the horizontal axis is the embeddings ranked from highest average correlation across characteristics (left) to lowest average correlation (right).  Note that S1 and G3 have a weak correlation across all embeddings, Collectively the Climate characteristics show the strongest correlation, followed by geomorphology/landcover. The soil and geology group represent the weakest correlation. This might be expected since soil , and geologic properties have high spatial variability as discussed earlier.

\subsection{Forward Modeling based evaluation}
\label{sec:evaluate}

In previous sections, we observed that KGSSL is able to recover characteristics under missing/uncertain scenarios. In this section, we take one step further and plug our retrieved values in state-of-the-art hydrological models to evaluate the gains achieved in streamflow prediction performance by these retrieved values compared to the missing/uncertain values.

LSTMs are extensively used for environmental modeling where both static and time-series variables are supplied as input (here, static characteristics are repeated at each time-step). However, the original RNN models were not designed to exploit static data. Recently, EA-LSTM ~\cite{kratzert2019towards} has emerged as one of the state-of-the-art ML-based forward models used in hydrology that processes the time-series meteorological drivers conditioned on static characteristics. Henceforth, we compare the streamflow prediction performance of the EA-LSTM model in two settings: KGSSL inferred features and original basin characteristics. We report Nash–Sutcliffe model efficiency coefficient (NSE) score for each forward model run.

\begin{table}[t]
    \centering
    \begin{tabular}{|c|c|}
        \hline
        \textbf{Method}     & \textbf{Mean NSE}\\
        \hline
        Baseline(uses known characteristics)                 & 0.704         \\
        Baseline (100\% missing)                & 0.491 \\
        \hline
        KGSSL (10\% missing)        & 0.697         \\
        KGSSL (50\% missing)        & 0.700         \\
        KGSSL (90\% missing)        & 0.664         \\
        KGSSL (100\% missing)   & 0.669         \\
        \hline
    \end{tabular}
    \caption{\small Model performance for different percentages of missing values. }
    \label{tab:MissingEALSTM}
    \vspace{-0.4in}
\end{table}

\subsubsection{Forward modeling with missing entity characteristics} \hfill\\
By design, KGSSL is trained using both supervised and unsupervised loss and has generalizations capability to infer the missing basin characteristics that can eventually enhance the streamflow prediction when basin characteristics are missing/not available. We train all models on all 531 basins during the train years and test the performance during the test years. Table~\ref{tab:MissingEALSTM} (first two rows) report performance of state-of-the-art EA-LSTM (baseline) trained with all and no static characteristics \cite{kratzert2019towards}. The baseline model where all characteristics are present performs $\mathbf{43\%}$ better (mean NSE) than the baseline model when some or all basin characters are missing. This shows the importance of the basin characteristics in modulating the driver-response network. 

To evaluate how much inferred basin features help in the forward model, we randomly treat 10\%, 50\%, 90\%, and 100\% of the characters to be missing. We impute missing characters using our KGSSL pipeline and run it through the forward model. Table~\ref{tab:MissingEALSTM} (last 4 rows) report NSE performance when our model was used to fill in the static characteristics for different percentages of missing values. We observe that the forward model trained with reconstructed characteristics from KGSSL with \textit{10\%} and \textit{50\%} missing values perform similar to the baseline trained with all characteristics. Further with \textit{90\%} missing characteristics, the forward model observes only 5\% drop when compared with baseline. In addition, for 100\% missing characteristics, we use a total unsupervised setting in our KGSSL framework, i.e., generate embeddings instead of characteristics. The KGSSL model with no supervision perform $\mathbf{35\%}$ better to the baseline with 100\% missing characteristics. Even more impressive is the fact that KGSSL with no supervision (last line) is only slightly worse than the baseline that uses known characteristics. We attribute this success to our framework's knowledge-guided component, which implicitly extracts complex correlations embedded in the input data. 

\begin{table}[t]
    \centering
    \begin{tabular}{|c|c|}
        \hline
        \textbf{Method}                             & \textbf{Mean NSE} \\
        \hline
        Baseline(actual characteristics)                                       & 0.560        \\
        Baseline ($0.5\sigma_i$ noise)  & 0.474	        \\
        Baseline  ($1\sigma_i$ noise)     & 0.245	        \\
        \hline
        KGSSL ($1$ year)                        & 0.460	      \\
        KGSSL ($2$ year)                        & 0.535	      \\
        KGSSL ($3$ year)                        & 0.554	      \\
        KGSSL ($9$ year)                        & 0.582	      \\
        \hline
    \end{tabular}
    \caption{\small Forward model performance with corrupted characteristics and using KGSSL embeddings. In KGSSL (n-year), the n refers to the number of years of data utilized to learn the embedding.}
    \label{tab:CorruptedEALSTM}
\end{table}

\begin{figure}[t]
    \centering
        \includegraphics[width=\linewidth]{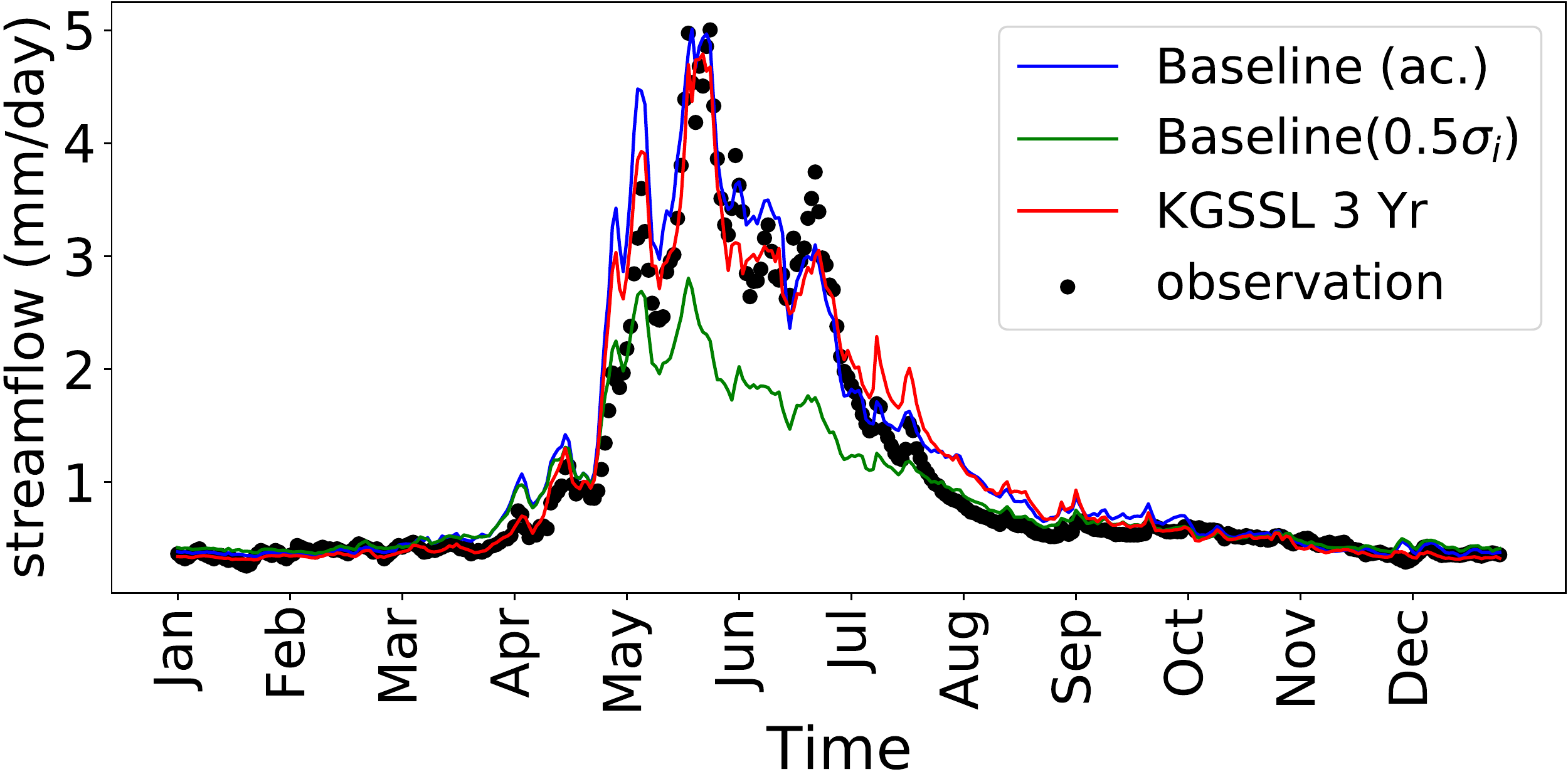}
       \vspace{-.2in} 
       \caption{\small Basin at year 1992 (Best seen in color)}
       \label{fig:streamflow}
       \vspace{-.25in} 
\end{figure}

\subsubsection{Forward modeling with corrupted entity characteristics}\hfill\\
As shown by Kratzert et al. ~\cite{kratzert2019towards}, uncertainty or corruption in basin-characteristics can have detrimental effect on the forward modeling. To demonstrate this, we trained the baseline model on the 400 train basins in the train years and test the performance on the 131 test basins in the test years. Table~\ref{tab:CorruptedEALSTM} (first row) shows the baseline performance of the forward model. To model uncertainty in the static characteristics, we add Gaussian noise ($\mathcal{N}(0,0.5\sigma_i)$,$\mathcal{N}(0,1\sigma_i)$) to test characteristics and measure the performance (Table~\ref{tab:CorruptedEALSTM} - $2^{nd}$, $3^{rd}$ row) of the baseline model. As expected, the forward model is susceptible to noise in the basin characteristic, and the performance drops significantly with slight noise. Specifically, the mean NSE drops by 50\% with a $1$ standard deviation noise. 

If the basin characteristics are corrupted, we can utilize representation obtained from the KGSSL trained in self-supervised manner using n-years of observations (note that this approach does not need any information about characteristics but it does need a small amount of data to create the embeddings). Table~\ref{tab:CorruptedEALSTM} (second set of rows), showcases the power of this methodology. As expected, the performance improves as we use more data to generate embeddings. Note that with only 2 years of data, the EALSTM model using \textit{KGSSL(2 year)} outperforms the EALSTM using \textit{Corrupted $\mathcal{N}(0,0.5\sigma_i)$} characteristics. Moreover, \textit{KGSSL(9 year)} generated with 9 years of train data outperform the model with \textit{actual characteristics}. 

In Figure \ref{fig:streamflow} we plot the actual observed streamflow (black dot) as well as the predicted streamflow using the various settings of the forward model. We observe that baseline imputation with corruption performs poorly and is nowhere close to the actual values. We also note that baseline prediction (blue line) closely matches our KGSSL predicted values using only 3 years of data (red line). We attribute this good result to our novel pretext task that is able to handle time varying physical processes.


\section{Discussion and Future Work}

In this work, we build a novel inverse framework KGSSL, and demonstrate the power of leveraging domain knowledge between entities in the context of streamflow. We performed extensive experiments on the hydrological benchmark dataset and show that KGSSL outperforms baseline significantly by a margin of 16-35 \% under various situations. KGSSL is a first-of-its-kind knowledge-guided framework that implicitly extracts system characteristics given its driver and response data. 

This paper addresses an important problem in the hydrologic domain, which is societally relevant. For much of the world, where water is kept in basins for later redistribution for societal needs, additional hydro climatological forecasting is required, particularly streamflow. Reasonable estimates are needed of the time patterns of streamflow about each basin and estimates of the release schedules necessary to meet societal contracts and environmental laws. As such, KGSSL can yield societally useful products for resource management and decision making.

We note that the proposed method is general and can add value in other applications such as computer vision (self-driving car), where additional features are used to capture variations in light \cite{dai2018dark}, weather \cite{volk2019towards}, and object poses \cite{alcorn2019strike}.Note that KGSSL learns representations without focusing on optimizing the response variable (i.e., streamflow prediction in our hydrology application). KGSSL  can be further extended by combining both the forward and inverse model in a unified framework that uses the inverse model to generate a representation and then uses the learned representation to modulate the forward model. Such an extension can leverage the recent work from task-aware modulation in machine-learning~\cite{vuorio2019multimodal,lin2021task,zintgraf2019fast}, and will be considered in future work. 


\section{Acknowledgement}
This work was funded by the NSF HDR Grant 1934721 and NSF FAI Grant 2147195. Access to computing facilities was provided by the Minnesota Supercomputing Institute. John Nieber's effort on this project was partially supported by the USDA National Institute of Food and Agriculture, Hatch/Multistate Project MN 12-109.


\bibliographystyle{ACM-Reference-Format}
\bibliography{main}


\appendix
\section{Appendix}

\subsection{Index of Catchment attributes used in our paper }

\begin{table}[H]
    \small
    \centering
    \vspace{-0.2in}
    \begin{tabular}{|c|c|c|}
        \hline
        Group     &index              & Name\\
        \hline
        Climate     &C1            & p mean \\       
                    &C2            & pet mean \\  
	                &C3            & p seasonality\\	    
	                &C4            & frac snow\\	     
	                &C5            & aridity\\	  
	                &C6            & high prec freq\\	   
	                &C7            & high prec dur\\	   
                    &C8        	& low prec freq\\
                    &C9        	& low prec dur\\
        \hline
        Soil geology	&S1            & carbonate rocks frac\\
                        &S2        	& geol permeability\\
                        &S3       	& soil depth pelletier\\
                        &S4        	& soil depth statsgo\\
                        &S5        	& soil porosity	\\
                        &S6        	& soil conductivity\\
                        &S7        	& max water content\\
                        &S8        	& sand frac	\\
                        &S9        	& silt frac	\\
                        &S10        	& clay frac\\
        \hline
        Geomorphology	&G1        & elev mean\\
                        &G2    	& slope mean\\
                        &G3   	& area gages2\\
                        &G4    	& frac forest\\
                        &G5    	& lai max\\
                        &G6    	& lai diff\\
                        &G7    	& gvf max\\
                        &G8    	& gvf diff\\
        \hline
    \end{tabular}
    \caption{Table of catchment characteristics used in this experiment, description of the characteristics is available and defined in \cite{Addor2017}}
    \label{tab:liststatic}
    \vspace{-0.3in}
\end{table}

\subsection{Hyperparameter Tuning}
We used grid search over a range of parameter values to find the hyperparameters, i.e., $\lambda_1$, $\lambda_2$, $\lambda_3$, batch\_size, the temperature of contrastive loss, learning rate, dimension of embedding.
Specifically, we considered the following possible parameter values.
\begin{itemize}
\item Dimension of embedding: 32, 64, 128,256
\item Learning\_rate : 0.0005, 0.001, 0.003, 0.005, 0.05
\item $\lambda_1$ :0.01, 0.1, 1, 10
\item $\lambda_2$ :1
\item $\lambda_3$ : 0.1, 1, 10
\item Batch\_size: 100, 200
\item Temp : 0.1, 0.5, 0.7, 1 
\end{itemize}

We trained our model on train basins in the training period and tested the model in the testing period on testing basins for each combination in the parameter set. We chose the parameter set with the least avg RMSE in the train basins during the validation years as the final parameter configuration

\subsection{Reproducibility}
CAMELS input data is freely available on the homepage of the NCAR at \url{https://ral.ucar.edu/solutions/products/camels}. The code is available at https://tinyurl.com/bdny7fk6 \href{https://tinyurl.com/bdny7fk6}{(KGSSL Code Link)}.

\end{document}